\documentclass{bmvc2k}

\usepackage{times}
\usepackage{epsfig}
\usepackage{graphicx}
\usepackage{amsmath}
\usepackage{amssymb}
\usepackage{soul}
\usepackage{paralist}

\title{In defense of OSVOS}

\addauthor{Yu Liu $^{~*}$ }{yu.liu@robotvision.org}{1}
\addauthor{Yutong Dai$^{~*}$}{
yutong.dai@adelaide.edu.au}{1}
\addauthor{Anh-Dzung Doan}{dung.doan@adelaide.edu.au}{1}
\addauthor{Lingqiao Liu}{lingqiao.liu@adelaide.edu.au}{1}
\addauthor{Ian Reid}{ian.reid@adelaide.edu.au}{1}

\addinstitution{School of Computer Science, \\ The University of Adelaide, AU
}

\runninghead{Video Segmentation}{In Defense of OSVOS}

\begin{document}

\maketitle

\begin{abstract}
As a milestone for video object segmentation, one-shot video object segmentation (OSVOS) has achieved a large margin compared to the conventional optical-flow based methods regarding to the segmentation accuracy. Its excellent performance mainly benefit from the three-step training mechanism, that are: (1) acquiring object features on the base dataset (i.e. ImageNet), (2) training the \textit{parent network} on the training set of the target dataset (i.e. DAVIS-2016) to be capable of differentiating the object of interest from the background.
(3) \textit{online fine-tuning} the interested object on 
the first frame of the target test set to overfit its appearance, then the model can be utilized to segment the same object in the rest frames of that video. In this paper, we argue that for the step (2), OSVOS has the limitation to `overemphasize' the generic semantic object information while `dilute' the instance cues of the object(s), which largely block the whole training process. Through adding a common module, video loss, which we formulate with various forms of constraints (including weighted BCE loss, high-dimensional triplet loss, as well as a novel mixed instance-aware video loss), to train the \textit{parent network} in the step (2), the network is then better prepared for the step (3), i.e. \textit{online fine-tuning} on the target instance.
Through extensive experiments using different network structures as the backbone, we show that the proposed video loss module can improve the segmentation performance significantly, compared to that of OSVOS. Meanwhile, since video loss is a common module, it can be generalized to other fine-tuning based methods and similar vision tasks such as depth estimation and saliency detection.
\end{abstract}

\section{Introduction}
\label{sec:intro}
With the popularity of all kinds of mobile device and sensors, countless video clips are uploaded and shared through the social media platforms and video websites every day. Smartly analysing these video clips are very useful yet quite challenging.
The revival of deep learning boosts the performance of many recognition tasks on static images to a level that can be matched with human beings, including object classification~\cite{liu2019learning,girshick2015fast,redmon2018yolov3}, semantic segmentation~\cite{chen2018deeplab, liu2018path,long2015fully} and object tracking~\cite{wang2018fast,cao2018openpose}. Compared to static images, video clips contain much more rich information, and the temporal correlations among inter-frame, if being used appropriately, it can significantly improve the performance of the corresponding tasks on static images. As one of the most active fields in computer vision community, video object segmentation aims to distinguish the foreground objects(s) from the background in pixel level. In 2017, one-shot video object segmentation (OSVOS) is proposed by Caelles et al ~\cite{caelles2017one}, as a milestone in this research field, which achieves over 10 \% improvements compared to the previous conventional methods regarding the segmentation accuracy. 

\begin{figure}[t]
\begin{center} 
 \includegraphics[width=1.0\linewidth]{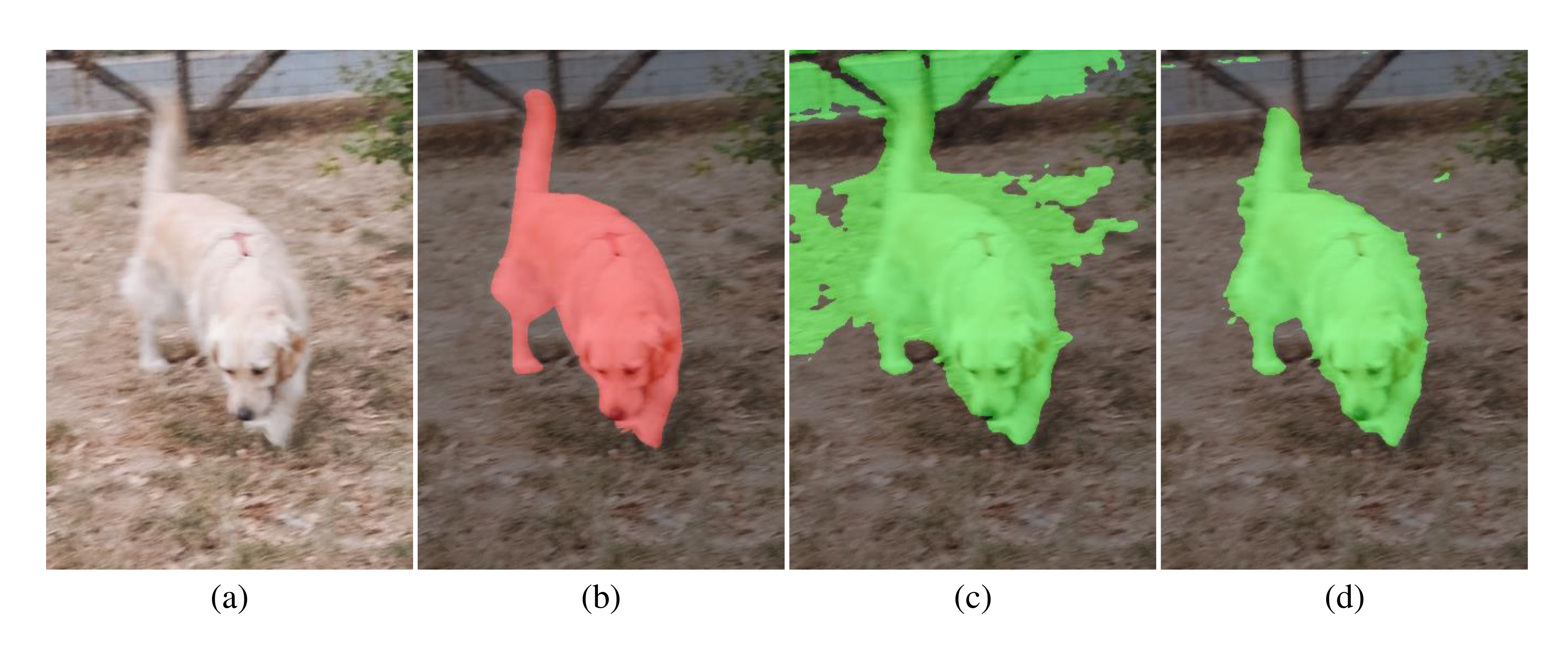}
\end{center}
   \caption{A visualized example of OSVOS and OSVOS-VL. (a) Image (b) Ground truth (c) Segmentation with OSVOS (d) Segmentation with OSVOS-VL (the proposed method with video loss).}
\end{figure}
\label{fig:dog}

\noindent
\paragraph{Motivation and principle of OSVOS}

The design of OSVOS is inspired by the perception process of human beings. Specifically, when we recognize an object, the first thing come into our view are the image features such as corners and textures of the scene, then we can distinguish the object(s) from background through shape and edge cues, which also named objectness. Finally, based on the rough localization from the above two steps, we will pay attention on the details of the target instance.   

In particular, OSVOS utilizes a fully-convolutional neural network (FCN)~\cite{long2015fully} to conduct video object segmentation, and the three phases are:
\begin{itemize}
\item \textbf{Acquire object features}: to acquire the generic semantic features from \textit{ImageNet}.

\item \textbf{Train parent network}: to train a network on DAVIS-2016 training set, which is capable of distinguishing the  foreground object(s) from the background.

\item \textbf{Online fine-tuning}: based on the \textit{ parent network}, to train the network which is overfitting the appearance of the target instance on the first frame.

\end{itemize}

\noindent
\paragraph{The Pros and Cons of OSVOS}
\vspace{-\topsep}
\begin{itemize}
	\setlength{\parskip}{0pt}
	\setlength{\itemsep}{0pt plus 1pt}
    \item \textbf{Pros:} The online fine-tuning process of OSVOS wishes to fully acquire the appearance of the target object in the first frame. Hence, it is capable of handling the fast moving, abrupt rotation as well as heavy occlusion, which are the limitations of the conventional optical flow based methods.
    \item \textbf{Cons:} \begin{inparaenum}[i)] \item when similar (noisy) objects appear in the subsequent frames of the video sequence, they will be wrongly segmented as foreground objects. \item when the appearance of the target object changes dramatically in the later phase of the video sequence, the algorithm fails to segment the new appearance parts. \end{inparaenum}

\end{itemize}

\paragraph{The motivation of video loss}
We propose the video loss in defense of OSVOS based on two observations:

\begin{itemize}
    \item For CNN, the low-level layers have relatively large spatial resolutions, and carry more details about the object instance, while the high-level layers have more stronger abstract and generalization ability, leading to carry more category information. Especially, in the second phase of OSVOS, i.e. training the \textit{parent network}, it actually tries to fine-tune the network to acquire the ability of distinguishing the objects from the background. However, it dilutes the `instance' information. And quickly adapts to the specific (target) instance, which is exactly the need of third phase (i.e. \textit{online finetuning}). Video loss can effectively `rectify' the training process of \textit{parent network}, and make it be better prepared for the \textit{online fine-tuning}.

    \item Each video is supposed to maintain an \textit{average object}, and through mapping, we expect that the objects from a same video are close to each other in the embedding space, while the objects from different videos are far away from each other. By this way, video loss can help the network to maintain an \textit{average object} for each video squence.
    
\end{itemize}

\section{Related Works}
\label{relatedworks}
For the task of semi-supervised video object segmentation, the annotations of first frame is given, and the algorithm is expected to infer the object(s) of interest in the rest frames of the video. According to design principle, the existing algorithms which achieve the state-of-the-art performance on DAVIS benchmark~\cite{Perazzi_2016_CVPR}
for semi-supervised video object segmentation can be roughly classified into three categories:

\subsection{Tracking based Methods}
In this category, one stream of methods employ the optical flow to track the mask from the previous frame to the current frame, including MSK~\cite{perazzi2017learning}, MPNVOS~\cite{sun2018mask} etc, one limitation of those methods is that they can not handle heavy occlusion and fast moving. Most recently, there are an emergence of methods which use the ReID technique to conduct the video object segmentation, including PReMVOS~\cite{luiten2018premvos} and FAVOS~\cite{cheng2018fast}. Specifically, FAVOS using ReID to tackle the part-based detection box first, and through merging the (box) region based segments to form the final segmentation. PReMVOS firstly generate instance segmentation proposals in each frame, and then take use of the ReID technique to do data association to pick the correct segments in temporal domain, which can largely reduce the background noises brought by other nearby or overlapped object(s).  

\subsection{Adaptation based Methods}
For this category of methods, the core idea is utilizing the mask priors acquired from the previous frame(s) to be the guidance, to supervise the prediction in the current frame. Specifically, Segflow~\cite{cheng2017segflow}
takes use of a parallel two-branch network to predict the segmentation as well as optical flow, through the bidirectional propagation between two frames, calculating optical flow and segmentation together and expecting them to benefit from each other. RGMP~\cite{wug2018fast} takes both annotations of the first frame and predicted mask of the previous frame as guidance, employs a Siamese encoder-decoder to conduct the mask propagation as well as detection, and with synthetic data to further boost the segmentation performance. OSMN~\cite{yang2018efficient} shares the similar design principle with RGMP, while the difference is that it uses an modulator to quickly adapt the first annotation to the previous frame, which can then be used by the segmentation network as the spatial prior.

\subsection{Fine-tuning based Methods}
Besides the aforementioned two categories of methods, there are some fine-tuning based methods which achieve the top performance in video object segmentation benchmark are OSVOS-S~\cite{maninis2017video}, OnVOS~\cite{voigtlaender2017online}, CINM~\cite{bao2018cnn} etc, and all of them are derived from OSVOS~\cite{caelles2017one}. Specifically, OSVOS-S~\cite{maninis2017video} aims to solve the problem of removing noisy object(s) with the help of instance segmentation. While OnAVOS~\cite{voigtlaender2017online} tries to enhance the network's ability for recognizing 
the new appearance of the target object(s) as well as suppressing the similar appearance carried by the noisy object(s).  CINM~\cite{bao2018cnn} is also initialized with the fine-tuning model, and employ a CNN to infer the markov random field (MRF) in spatial domain, and with optical flow to track the segmented object(s) in temporal domain.

\subsection{Video Loss}
In this paper, targeting to improve the fine-tuning methods, with OSVOS as an entry, we deliver a tiny head \textit{video loss}. As aforementioned, our observation is based on the `delayed' learning process for target instance(s) between the \textit{parent network} and \textit{online fine-tuning}. Through incorporating 
a basic component, i.e. video loss, we achieve better performance regarding to segmentation accuracy compared to OSVOS using exactly same backbone network structure(s). Furthermore, considering that it may not always be easy to distinguish the object(s) from background in 2D image coordinate, we further utilize metric learning and the proposed mixed instance-aware video loss to enforce the pixels, after mapping through a FCN in high-dimensional space, which belong to target object(s) or background are supposed to be closed with each other, while any two pixels with one belong to the target object(s) and the other belong to background are supposed to has a relatively far distance with each other. Through employing the proposed \textit{video losses}, the performance has been significantly improved regarding to the segmentation accuracy, and some noisy objects have been effectively removed. Moreover, since \textit{video loss} is a common building block, it can be generalized to all kinds of fine-tuning based methods including, but not limit to OnVOS~\cite{voigtlaender2017online}, OSVOS-S~\cite{maninis2017video}, CINM~\cite{bao2018cnn} etc.

\begin{figure*}[t]
\begin{center}
 \includegraphics[width=1.0\linewidth]{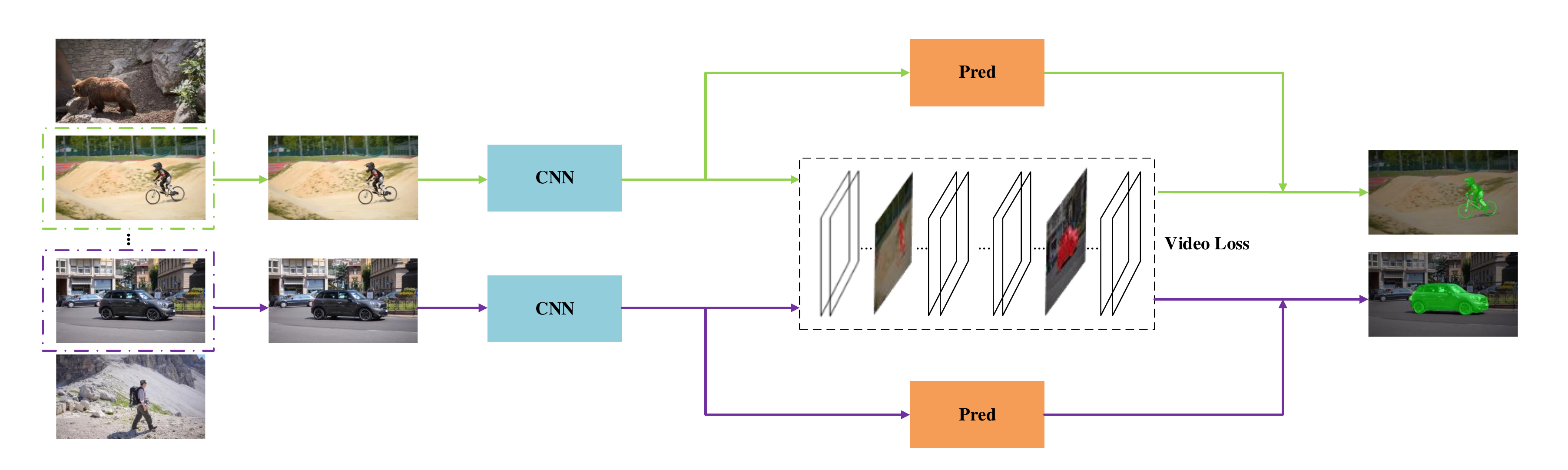}
\end{center}
   \caption{The workflow of OSVOS-VL. Compared to OSVOS, only a tiny head, video loss block, is added.}
\label{fig:workflow}   
\end{figure*}

\section{Methodology}
The motivation, design and key implementation of \textit{video loss} will be illustrated in detail.


\subsection{Overview}
The assumption for \textit{video loss} is that, different objects are linear separable in high-dimentional space, i.e. the feature space. Meanwhile, the euclidean distances of the features of the same object are supposed to be smaller than that of different objects. The workflow of OSVOS-VL is shown in Figure~\ref{fig:workflow}. As can be seen, \textit{video loss} just like a light-weight head being parallel with the prediction part, thus the extra time cost is insignificant. Once the better features are obtained after training  of \textit{parent network}, it would ease the learning processing of \textit{online fine-tuning} stage and is much prone to achieve accurate segmentation results compared to that of OSVOS. This design can be viewed as maintaining an \textit{average target object} for each video, and expecting the objects from different videos are much more far away than that of from the same video, which effectively prevents the background noise from other objects (of no interest). 
We deliver three types of video loss (VL) in this paper. The first one is the two dimensional video loss (2D-VL), which make the \textit{parent network} to push away different objects in image coordinates. The second and third are the high-dimensional video loss (HD-VL). Established on 2D-VL, the HD-VL further maps 2D features to high dimensional space, and clusters pixels which belong to the same instance together, and utilize object centers in HD-space as constraints.  

\subsection{Two Dimensional Video Loss}
\begin{figure*}[t]
\begin{center}
 \includegraphics[width=0.7\linewidth]{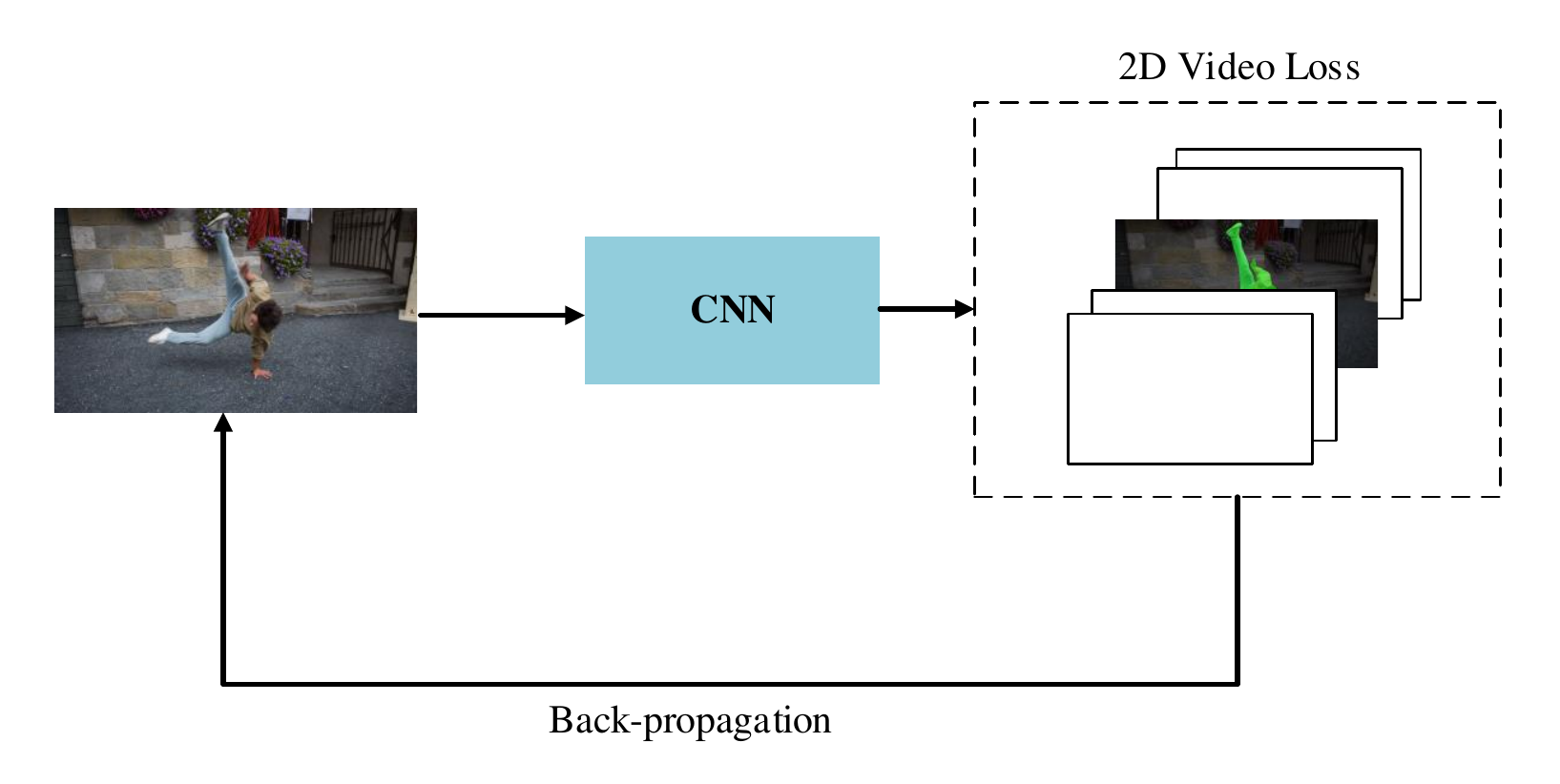}
\end{center}
   \caption{The illustration of two-dimensional video loss.}
\label{fig:2d_videoloss} 
\end{figure*}

In OSVOS~\cite{caelles2017one}, considering the sample imbalance between the target object(s) and the background, weight cross entropy loss is employed to conduct the pixel-wise segmentation task. The expression of weighted cross entropy loss as follows:

\begin{equation}
    \begin{aligned}
        { L }_{ 2d }=-\frac { { Y }_{ - } }{ { Y } } \sum _{ j\in { Y }_{ + } }^{  }{ \log { P\left( { y }_{ j }=1|X \right)  }  }-\left( 1-\frac { { Y }_{ - } }{ { Y } }  \right) \sum _{ j\in { Y }_{ - } }^{  }{ \log { P\left( { y }_{ j }=0|X \right)  }  } 
    \end{aligned}
    \label{eq1}
\end{equation}
Where $X$ is the input image, $y_i \in 0,1,j=1,\dots,|X|$ is the pixelwise binary label of $X$, and $Y$ and $Y_-$ are the positive and negative labeled pixels. $P(\cdot)$ is obtained by applying a sigmoid to the activation of the final layer. $|Y_-|/|Y|$ is employed for the purpose of training the imbalanced binary task as in~\cite{xie2015holistically}.

OSVOS~\cite{caelles2017one} only rely on weighted cross entropy loss to fine-tune \textit{parent network}, but we argue that it will mix up all of the objectness features in DAVIS dataset, without classifying which kind of objects the foreground belongs to. It may make the online fine-tuning process more harder to recognize which instance the object is. Therefore, we propose 2D-VL to force the network to learn features of different instances during the training of \textit{parent network}.  

For this purpose, we add the identity of each video ($v_{id}$) into the training process as input. After recognizing the $v_{id}$ of the training (image) data, the network can update each specific (video) category through back propagation. Please note, 2D-VL share the same expression with Equation~\ref{eq1},
but different from the prediction branch, our 2D-VL only updates corresponding (video) category directly, as illustrated in Figure~\ref{fig:2d_videoloss}.

\subsection{High Dimensional Video Loss}
\begin{figure*}[t]
\begin{center}
 \includegraphics[width=0.8\linewidth]{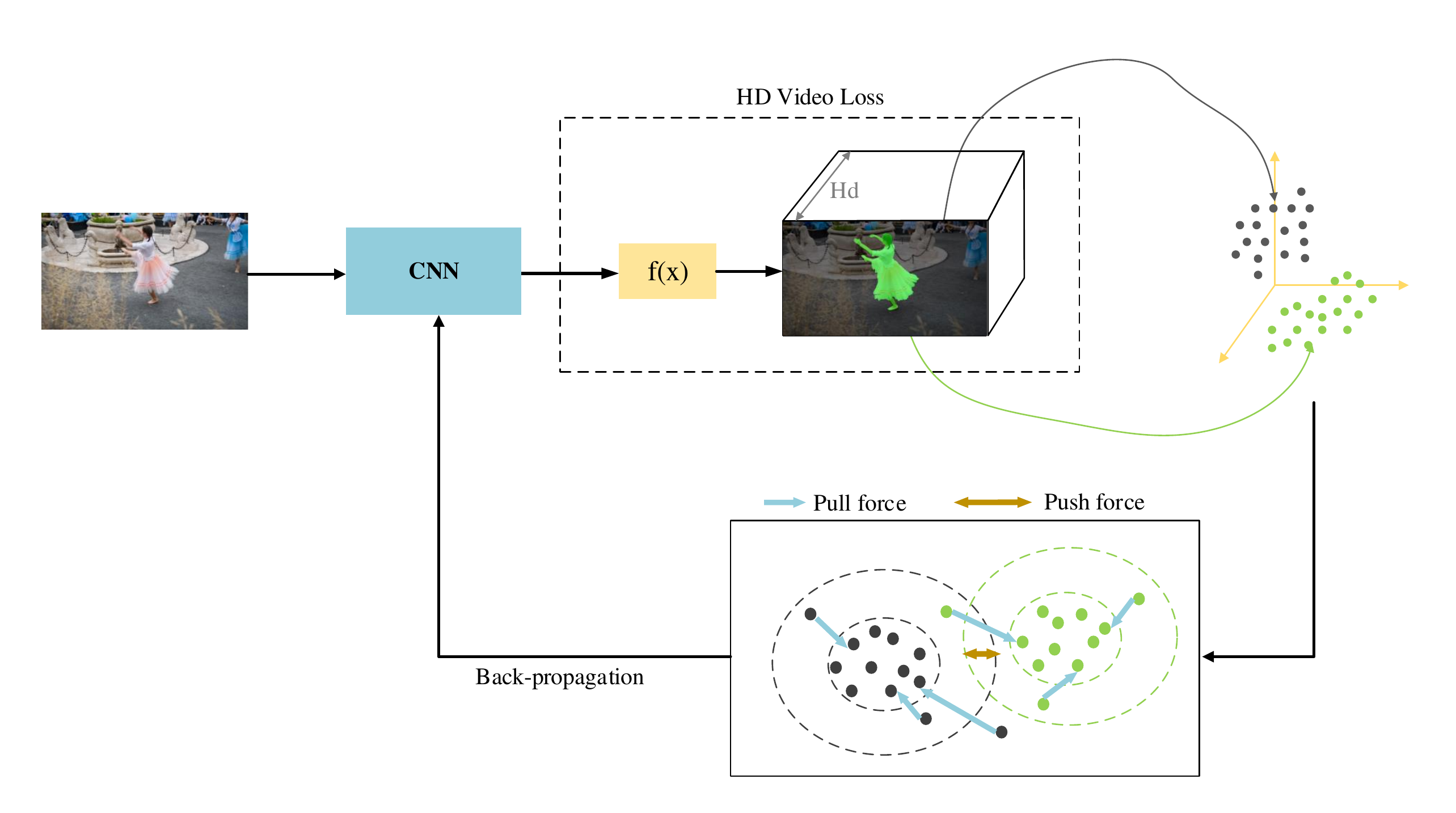}
\end{center}
   \caption{The illustration of high-dimensional video loss.}
\label{fig:hd_videoloss}   
\end{figure*}

With the observation that objects in 2D dimension usually have similar appearances or shapes, which brings too much confusion to the network to distinguish an object accurately, we propose to map the prediction to a high-dimensional space firstly, and expecting that after mapping, the distance among different objects is enlarged. The mapping process in high dimensional space is shown in Fig.~\ref{fig:hd_videoloss}. In HD space, we look forward to clustering embeddings from different objects into different groups. 

In PML~\cite{chen2018blazingly}, a modified triplet loss is utlized to pull samples with same identity close to each other, and only constrain the smallest negative points and smallest positive points. Inspried but different from that, we randomly sample 256 points in both foreground parts and background parts, pulling points from the same part together and push points from different parts away. The triplet loss is defined as
\begin{equation}
    \begin{aligned}
    { L }_{ hd\_ tl }=y\left\| f\left( { x }_{ 1 } \right) -f\left( { x }_{ 2 } \right)  \right\|+
    \left( 1-y \right) max\left( 0,\quad \lambda -\left\| f\left( { x }_{ 1 } \right) -f\left( { x }_{ 2 } \right)  \right\|  \right) 
\end{aligned}
\end{equation}
where $y=1$ when point $x_{1}$, $x_{2}$ belong to a same cluster (foreground or background) and $y=0$ when point $x_{1}$, $x_{2}$ belong to different clusters.

\subsection{Mixed Instance-aware Video Loss }

\noindent
\textbf{Contrastive Center Loss}
Inspired by the work~\cite{de2017semantic}, we define a contrastive center loss for the purpose of pulling embeddings with the same label close to each other while pushing embeddings with different labels far away from each other. To restrict the entire foreground area and background area, we first calculate the center point of each part. Then we use the contrastive center loss function to penalize the distance between these two center points. The motivation behind this is to restrict distribution of foreground embeddings and reduce the amount of computation.
\begin{equation}
    { L }_{ hd\_ cl }=max\left( 0,\quad \lambda -\left\| { \mu  }_{ + }-{ \mu  }_{ - } \right\|  \right) 
\end{equation}
where ${ \mu  }_{ + }$ represents the center point of foreground cluster, and ${ \mu  }_{ - }$ represents the center point of background cluster, both in high-dimensional space.

\noindent
\textbf{Mixed Loss}
Contrastive center loss is a loss restricting the overall distribution of examples, while triplet loss considers the constraints in pixel level. In order to combine two types of constraints, here we define a mixed loss as
\begin{equation}
    { L }_{ hd\_ mix }={ \beta  }_{ 1 }{ L }_{ hd\_ cl }+{ \beta  }_{ 2 }{ L }_{ hd\_ tl }
\end{equation}

\noindent Where ${\beta}_1$  and ${\beta}_2$ are the coefficients for balancing two loss terms.


\subsection{Training}
In order to form a fair comparison with OSVOS~\cite{caelles2017one}, 
we adopt the same settings for the training of \textit{parent network} and \textit{online fine-tuning} except the training epochs. Specifically, SGD solver with momentum 0.9 is used, learning rate is 1e-8, the weight decay is 5e-4. Batch size is 1 for VGG16 based experiments and is 2 for MobileNet~\cite{howard2017mobilenets} based experiments, respectively. For training the \textit{parent network}, fine-tuning of 240 epochs is conducted based on the initialization of ImageNet~\cite{deng2009imagenet} features. For online fine-tuning, 10k iterations of fine-tuning is applied for all of the experiments for the fair comparison.

\section{Experimental Results}

\begin{table*}[t]
\begin{center}
\begin{tabular}{c|c|c|c}

Method & Parent Network & Fine\-tuning & Backbone \\
\hline
OSVOS & \textbf{52.5} & 75.0 & VGG16
\\

OSVOS-V2d & 50.8 & \textbf{76.2} & VGG16
\\


\hline
OSVOS & 53.1 & 65.7 & MobileNet
\\

OSVOS-V2d & \textbf{54.1} & \textbf{66.2} & MobileNet
\\



\end{tabular}

\end{center}
\caption{J Mean of OSVOS and OSVOS-V2d with different backbone network structures.}
\label{davis2016_number}
\end{table*}

\begin{table*}[t]
\begin{center}
\begin{tabular}{c|c|c}

Method & Parent Network & Fine\-tuning  \\
\hline
OSVOS & 53.1 & 65.7
\\

OSVOS-V2d & 54.1 & 66.2 
\\

OSVOS-Vhd & 53.7 & 66.9
\\

OSVOS-Vmixed & \textbf{58.6} & \textbf{67.5}
\\

\end{tabular}

\end{center}
\caption{J Mean of OSVOS, OSVOS-V2d, OSVOS-Vhd, OSVOS-Vmixed. With MobileNet as backbone, and 20 is dimensioins for the embedding of OSVOS-Vhd and OSVOS-Vmixed.}
\label{loss_mobilenet}
\end{table*}

\subsection{Dataset}
DAVIS-2016~\cite{Perazzi_2016_CVPR} is the most widely used dataset for video object segmentation, which is composed of 50 videos with pixel-wise annotations for single-object. Among them, 30 video sequences are chosen as training set, and the other 20 video sequences are utilized as test set.

\subsection{Quantitative Results}
In Table~\ref{davis2016_number}, J Mean for both \textit{parent network} and \textit{online fine-tuning} with different structures as backbones are listed out. As can be seen, for both of the experiments which based on VGG16 and MobileNet, OSVOS+ video loss achieve the better performance during \textit{onlie fine-tuning} phase, while with comparable performance with OSVOS during \textit{parent network} training phase, which proves our assumption that video loss, as a common module, is effective in helping the (FCN) network to recognize the target instance.
In Table~\ref{loss_mobilenet}, compared to OSVOS with video loss utilized in 2D (OSVOS-V2d), OSVOS with high-dimensional loss (OSVOS-Vhd, OSVOS-Vmixed) performs better, which is matched with our observation that sometimes it is much more easier for similar features to be distinguished in high-dimensional space than that of in two-dimensional space. Please note, all of our experiments trained with 10k iterations and without any post-processing for the purpose of fair comparison and saving training time, which is slightly different from ~\cite{caelles2017one}, and the preliminary experiments we tested show that as the training iterations increasing (around 20k iterations), which can replicate the numbers that the paper~\cite{caelles2017one} report.

\subsection{Qualitative Results}
\begin{figure*}[t]
\begin{center}
 \includegraphics[width=1.0\linewidth]{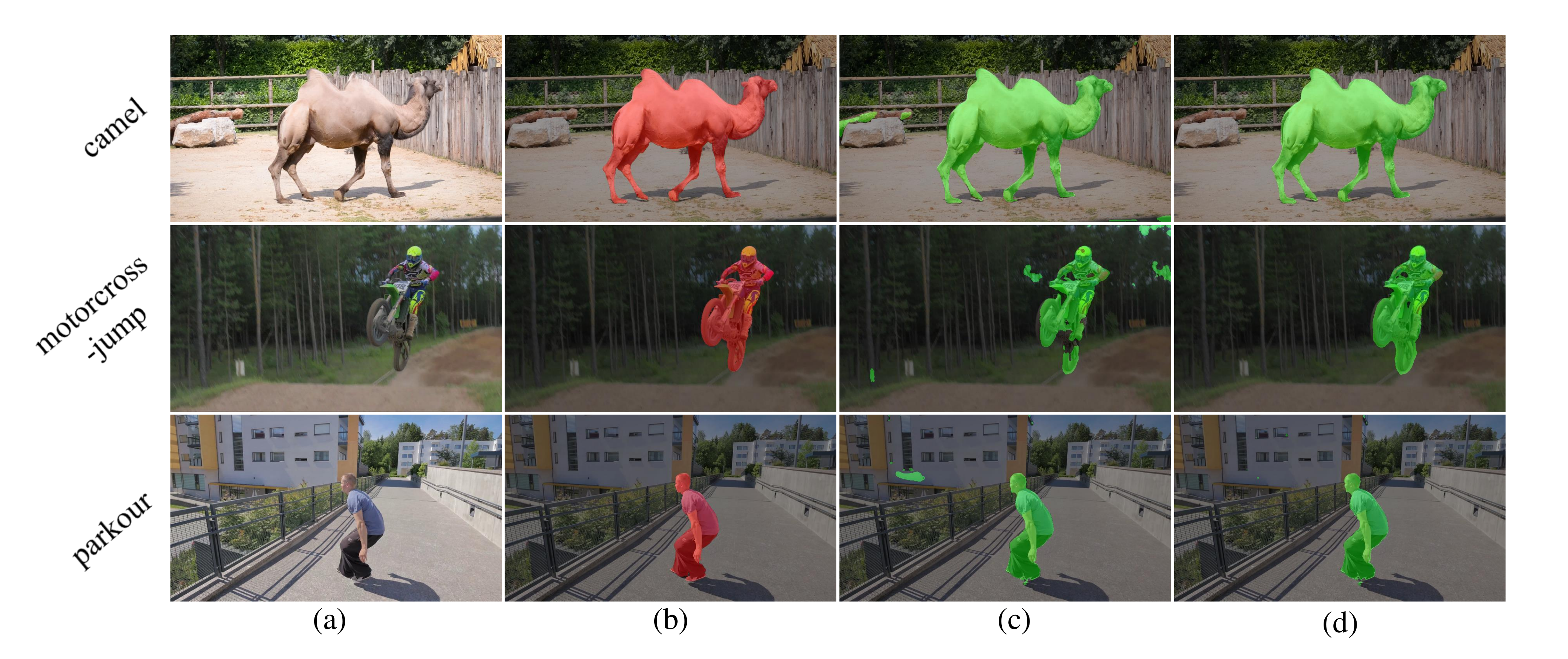}
\end{center}
   \caption{Visualization results of OSVOS and OSVOS-VL. (a) Input (b) Ground Truth (c) Segmentation from OSVOS (d) Segmentation from OSVOS-V2d.}
\label{vis_vl}   
\end{figure*}

We also provide some visualized comparisons in Figure~\ref{vis_vl} between OSVOS and OSVOS-V2d, as can be seen, among the results acquired by OSVOS, wrong segments are accompanied in the sourrondings of the target object, we suspect that is because only rely on prediction loss in OSVOS can not distinguish instance information between the target object and background (noisy) objects. In contrast, OSVOS-VL effectively remove the noisy parts compard to that of OSVOS.

\subsection{Performance on per sequence}
In order to have a better understanding of the work principle of the proposed video loss block, we illustrate performance comparsion of OSVOS and OSVOS-V2d on per sequence, as shown in Table ~\ref{sequence_number}, for both VGG16 and MobileNet based experiments, in 12 out of 20 sequences, OSVOS-V2d achieve better performance than OSVOS, in some sequences such as \textit{bmx-trees}, \textit{camel}, \textit{dance-twirl}, \textit{dog}, \textit{drift-chicane}, \textit{drift-straight}, \textit{motocross-jump}, \textit{paragliding-launch}, OSVOS-V2d achieves consistent improvements on both backbones, and these sequences usually contain abrupt motions or noisy objects which share the similar appearance with the target object.

\begin{table*}[!t]\small
\begin{center}
\scalebox{0.9}{
\begin{tabular}{c|c|c|c|c}

    Sequence & VGG+OSVOS & VGG+OSVOS-V2d & MN+OSVOS & MN+OSVOS-V2d 
    \\
    \hline
    Blackswan & \textbf{94.1} & 93.6 & 93.3 & \textbf{94.0} \\
    bmx-trees & 52.8 & \textbf{58.5} & 42.0 & \textbf{42.8} \\
    breakdance & 67.6 & \textbf{67.7} & \textbf{75.1} & 70.0 \\
    camel & 83.7 & \textbf{85.8} & 70.2 & \textbf{75.1} \\
    car-roundabout & \textbf{88.3} & 75.6 & \textbf{83.0} & 74.2 \\
    car-shadow & \textbf{88.6} & 83.5 & \textbf{74.3} & 74.2 \\
    cows & \textbf{95.2} & 94.9 & \textbf{90.7} & 87.9 \\
    dance-twirl & 60.7 & \textbf{64.9} & 63.4 & \textbf{66.2} \\
    dog & 72.6 & \textbf{88.3} & 88.7 & \textbf{90.8} \\
    drift-chicane & 61.3 & \textbf{73.9} & 26.9 & \textbf{36.0} \\
    drift-straight & 56.4 & \textbf{61.8} & 35.5 & \textbf{35.7} \\
    goat & \textbf{88.1} & 87.9 & 82.0 & \textbf{85.5} \\
    horsejump-high & \textbf{84.5} & 81.5 & \textbf{69.8} & 69.1 \\
    kite-surf & \textbf{75.3} & 73.9 & 54.9 & \textbf{55.3} \\
    libby & 75.4 & \textbf{77.2} & \textbf{69.4} & 68.3 \\
    motocross-jump & 60.2 & \textbf{67.3} & 49.6 & \textbf{52.7} \\
    paragliding-launch & 63.9 & \textbf{64.0} & 56.3 & \textbf{58.5} \\
    parkour & 89.0 & \textbf{89.2} & \textbf{81.5} & 73.6 \\
    scooter-black & \textbf{58.2} & 35.4 & 57.6 & \textbf{62.3} \\
    soapbox & 84.3 & \textbf{86.1} & \textbf{49.5} & 46.3 \\
    \hline
\end{tabular}
}
\end{center}
\caption{J Mean of OSVOS and OSVOS-V2d on per sequence. VGG denotes VGG16 and MN denotes MobileNet as the feature extractor.}
\label{sequence_number}
\end{table*}

\section{Conclusion}
In this paper, we deliver a common module, video loss, for video object segmentation, which is tailored to overcome the limitation of fine-tuning methods, during the phase of training \textit{parent network}, dilute the instance information, hence delay the overall training process. Considering in CNN, the shallow layers usually contain much rich details of object(s) which are the key cues to specify different instances, while the deeper layers have more stronger generalization ability to recognize generic objects. Various video losses are proposed as the constraints to supervise the training process of \textit{parent network}, which is effective in removing the noisy objects. Once the training process is finished, the \textit{parent network} is well prepared to adapt to the instance quickly during \textit{online fine-tuning}. One of our future interests will be extending the video loss into other fine-tuning methods such OSVOS-S, OnVOS. Another one will be with the help of the network search technique to automatically decide the training epochs and learning rate.

\bibliography{egbib}

\begin{thebibliography}{25}
\providecommand{\natexlab}[1]{#1}
\providecommand{\url}[1]{\texttt{#1}}
\expandafter\ifx\csname urlstyle\endcsname\relax
  \providecommand{\doi}[1]{doi: #1}\else
  \providecommand{\doi}{doi: \begingroup \urlstyle{rm}\Url}\fi

\bibitem[Bao et~al.(2018)Bao, Wu, and Liu]{bao2018cnn}
Linchao Bao, Baoyuan Wu, and Wei Liu.
\newblock Cnn in mrf: Video object segmentation via inference in a cnn-based
  higher-order spatio-temporal mrf.
\newblock In \emph{Proceedings of the IEEE Conference on Computer Vision and
  Pattern Recognition}, pages 5977--5986, 2018.

\bibitem[Caelles et~al.(2017)Caelles, Maninis, Pont-Tuset, Leal-Taix{\'e},
  Cremers, and Van~Gool]{caelles2017one}
Sergi Caelles, Kevis-Kokitsi Maninis, Jordi Pont-Tuset, Laura Leal-Taix{\'e},
  Daniel Cremers, and Luc Van~Gool.
\newblock One-shot video object segmentation.
\newblock In \emph{Proceedings of the IEEE conference on computer vision and
  pattern recognition}, pages 221--230, 2017.

\bibitem[Cao et~al.(2018)Cao, Hidalgo, Simon, Wei, and Sheikh]{cao2018openpose}
Zhe Cao, Gines Hidalgo, Tomas Simon, Shih-En Wei, and Yaser Sheikh.
\newblock Openpose: realtime multi-person 2d pose estimation using part
  affinity fields.
\newblock \emph{arXiv preprint arXiv:1812.08008}, 2018.

\bibitem[Chen et~al.(2018{\natexlab{a}})Chen, Papandreou, Kokkinos, Murphy, and
  Yuille]{chen2018deeplab}
Liang-Chieh Chen, George Papandreou, Iasonas Kokkinos, Kevin Murphy, and Alan~L
  Yuille.
\newblock Deeplab: Semantic image segmentation with deep convolutional nets,
  atrous convolution, and fully connected crfs.
\newblock \emph{IEEE transactions on pattern analysis and machine
  intelligence}, 40\penalty0 (4):\penalty0 834--848, 2018{\natexlab{a}}.

\bibitem[Chen et~al.(2018{\natexlab{b}})Chen, Pont-Tuset, Montes, and
  Van~Gool]{chen2018blazingly}
Yuhua Chen, Jordi Pont-Tuset, Alberto Montes, and Luc Van~Gool.
\newblock Blazingly fast video object segmentation with pixel-wise metric
  learning.
\newblock In \emph{Proceedings of the IEEE Conference on Computer Vision and
  Pattern Recognition}, pages 1189--1198, 2018{\natexlab{b}}.

\bibitem[Cheng et~al.(2017)Cheng, Tsai, Wang, and Yang]{cheng2017segflow}
Jingchun Cheng, Yi-Hsuan Tsai, Shengjin Wang, and Ming-Hsuan Yang.
\newblock Segflow: Joint learning for video object segmentation and optical
  flow.
\newblock In \emph{Proceedings of the IEEE international conference on computer
  vision}, pages 686--695, 2017.

\bibitem[Cheng et~al.(2018)Cheng, Tsai, Hung, Wang, and Yang]{cheng2018fast}
Jingchun Cheng, Yi-Hsuan Tsai, Wei-Chih Hung, Shengjin Wang, and Ming-Hsuan
  Yang.
\newblock Fast and accurate online video object segmentation via tracking
  parts.
\newblock In \emph{Proceedings of the IEEE Conference on Computer Vision and
  Pattern Recognition}, pages 7415--7424, 2018.

\bibitem[De~Brabandere et~al.(2017)De~Brabandere, Neven, and
  Van~Gool]{de2017semantic}
Bert De~Brabandere, Davy Neven, and Luc Van~Gool.
\newblock Semantic instance segmentation with a discriminative loss function.
\newblock \emph{arXiv preprint arXiv:1708.02551}, 2017.

\bibitem[Deng et~al.(2009)Deng, Dong, Socher, Li, Li, and
  Fei-Fei]{deng2009imagenet}
Jia Deng, Wei Dong, Richard Socher, Li-Jia Li, Kai Li, and Li~Fei-Fei.
\newblock Imagenet: A large-scale hierarchical image database.
\newblock In \emph{2009 IEEE conference on computer vision and pattern
  recognition}, pages 248--255. Ieee, 2009.

\bibitem[Girshick(2015)]{girshick2015fast}
Ross Girshick.
\newblock Fast r-cnn.
\newblock In \emph{Proceedings of the IEEE international conference on computer
  vision}, pages 1440--1448, 2015.

\bibitem[Howard et~al.(2017)Howard, Zhu, Chen, Kalenichenko, Wang, Weyand,
  Andreetto, and Adam]{howard2017mobilenets}
Andrew~G Howard, Menglong Zhu, Bo~Chen, Dmitry Kalenichenko, Weijun Wang,
  Tobias Weyand, Marco Andreetto, and Hartwig Adam.
\newblock Mobilenets: Efficient convolutional neural networks for mobile vision
  applications.
\newblock \emph{arXiv preprint arXiv:1704.04861}, 2017.

\bibitem[Liu et~al.(2018)Liu, Qi, Qin, Shi, and Jia]{liu2018path}
Shu Liu, Lu~Qi, Haifang Qin, Jianping Shi, and Jiaya Jia.
\newblock Path aggregation network for instance segmentation.
\newblock In \emph{Proceedings of the IEEE Conference on Computer Vision and
  Pattern Recognition}, pages 8759--8768, 2018.

\bibitem[Liu et~al.(2019)Liu, Liu, Rezatofighi, Do, Shi, and
  Reid]{liu2019learning}
Yu~Liu, Lingqiao Liu, Hamid Rezatofighi, Thanh-Toan Do, Qinfeng Shi, and Ian
  Reid.
\newblock Learning pairwise relationship for multi-object detection in crowded
  scenes.
\newblock \emph{arXiv preprint arXiv:1901.03796}, 2019.

\bibitem[Long et~al.(2015)Long, Shelhamer, and Darrell]{long2015fully}
Jonathan Long, Evan Shelhamer, and Trevor Darrell.
\newblock Fully convolutional networks for semantic segmentation.
\newblock In \emph{Proceedings of the IEEE conference on computer vision and
  pattern recognition}, pages 3431--3440, 2015.

\bibitem[Luiten et~al.(2018)Luiten, Voigtlaender, and Leibe]{luiten2018premvos}
Jonathon Luiten, Paul Voigtlaender, and Bastian Leibe.
\newblock Premvos: Proposal-generation, refinement and merging for video object
  segmentation.
\newblock \emph{arXiv preprint arXiv:1807.09190}, 2018.

\bibitem[Maninis et~al.(2017)Maninis, Caelles, Chen, Pont-Tuset,
  Leal-Taix{\'e}, Cremers, and Van~Gool]{maninis2017video}
Kevis-Kokitsi Maninis, Sergi Caelles, Yuhua Chen, Jordi Pont-Tuset, Laura
  Leal-Taix{\'e}, Daniel Cremers, and Luc Van~Gool.
\newblock Video object segmentation without temporal information.
\newblock \emph{arXiv preprint arXiv:1709.06031}, 2017.

\bibitem[Perazzi et~al.(2016)Perazzi, Pont-Tuset, McWilliams, Van~Gool, Gross,
  and Sorkine-Hornung]{Perazzi_2016_CVPR}
Federico Perazzi, Jordi Pont-Tuset, Brian McWilliams, Luc Van~Gool, Markus
  Gross, and Alexander Sorkine-Hornung.
\newblock A benchmark dataset and evaluation methodology for video object
  segmentation.
\newblock In \emph{The IEEE Conference on Computer Vision and Pattern
  Recognition (CVPR)}, June 2016.

\bibitem[Perazzi et~al.(2017)Perazzi, Khoreva, Benenson, Schiele, and
  Sorkine-Hornung]{perazzi2017learning}
Federico Perazzi, Anna Khoreva, Rodrigo Benenson, Bernt Schiele, and Alexander
  Sorkine-Hornung.
\newblock Learning video object segmentation from static images.
\newblock In \emph{Proceedings of the IEEE Conference on Computer Vision and
  Pattern Recognition}, pages 2663--2672, 2017.

\bibitem[Redmon and Farhadi(2018)]{redmon2018yolov3}
Joseph Redmon and Ali Farhadi.
\newblock Yolov3: An incremental improvement.
\newblock \emph{arXiv preprint arXiv:1804.02767}, 2018.

\bibitem[Sun et~al.(2018)Sun, Yu, Li, and Wang]{sun2018mask}
Jia Sun, Dongdong Yu, Yinghong Li, and Changhu Wang.
\newblock Mask propagation network for video object segmentation.
\newblock \emph{arXiv preprint arXiv:1810.10289}, 2018.

\bibitem[Voigtlaender and Leibe(2017)]{voigtlaender2017online}
Paul Voigtlaender and Bastian Leibe.
\newblock Online adaptation of convolutional neural networks for video object
  segmentation.
\newblock \emph{arXiv preprint arXiv:1706.09364}, 2017.

\bibitem[Wang et~al.(2018)Wang, Zhang, Bertinetto, Hu, and Torr]{wang2018fast}
Qiang Wang, Li~Zhang, Luca Bertinetto, Weiming Hu, and Philip~HS Torr.
\newblock Fast online object tracking and segmentation: A unifying approach.
\newblock \emph{arXiv preprint arXiv:1812.05050}, 2018.

\bibitem[Wug~Oh et~al.(2018)Wug~Oh, Lee, Sunkavalli, and Joo~Kim]{wug2018fast}
Seoung Wug~Oh, Joon-Young Lee, Kalyan Sunkavalli, and Seon Joo~Kim.
\newblock Fast video object segmentation by reference-guided mask propagation.
\newblock In \emph{Proceedings of the IEEE Conference on Computer Vision and
  Pattern Recognition}, pages 7376--7385, 2018.

\bibitem[Xie and Tu(2015)]{xie2015holistically}
Saining Xie and Zhuowen Tu.
\newblock Holistically-nested edge detection.
\newblock In \emph{Proceedings of the IEEE international conference on computer
  vision}, pages 1395--1403, 2015.

\bibitem[Yang et~al.(2018)Yang, Wang, Xiong, Yang, and
  Katsaggelos]{yang2018efficient}
Linjie Yang, Yanran Wang, Xuehan Xiong, Jianchao Yang, and Aggelos~K
  Katsaggelos.
\newblock Efficient video object segmentation via network modulation.
\newblock In \emph{Proceedings of the IEEE Conference on Computer Vision and
  Pattern Recognition}, pages 6499--6507, 2018.

\end{thebibliography}
\end{document}